\documentclass[pdflatex,sn-mathphys-num]{sn-jnl}

\usepackage{graphicx}%
\usepackage{multirow}%
\usepackage{amsmath,amssymb,amsfonts}%
\usepackage{amsthm}%
\usepackage{mathrsfs}%
\usepackage[title]{appendix}%
\usepackage{xcolor}%
\usepackage{textcomp}%
\usepackage{manyfoot}%
\usepackage{booktabs}%
\usepackage{algorithm}%
\usepackage{algorithmicx}%
\usepackage{algpseudocode}%
\usepackage{listings}%

\theoremstyle{thmstyleone}%
\theoremstyle{thmstyletwo}%

\theoremstyle{thmstylethree}%

\begin{document}

\title[Learning to Rank Propagators in Constraint Programming]{Learning to Rank Propagators in Constraint Programming}


\author*[1]{\fnm{Matija} \sur{Lojovi\'c}}\email{matija.lojovic@matf.bg.ac.rs}

\author[1]{\fnm{Mladen} \sur{Nikoli\'c}}\email{mladen.nikolic@matf.bg.ac.rs}

\affil*[1]{\orgdiv{Faculty of Mathematics}, \orgname{University of Belgrade}, \orgaddress{\street{Studentski trg 16}, \city{Belgrade}, \postcode{11000}, \country{Serbia}}}


\abstract{The order in which propagators are executed can substantially affect the performance of constraint programming solvers, yet propagator priorities are typically determined by fixed, manually designed heuristics. In this work, we explore learning instance-dependent propagator priorities from static problem features. 
    We formulate the task as a learning-to-rank problem, addressing the challenge that the quality of a priority assignment can be measured through solver runtime, while the desired priorities of individual propagators are not directly observable.
    A self-attention-based model jointly scores the propagators present in an instance and is trained using pairwise preferences derived from solver runtimes measured under randomized propagator priority assignments, with a Plackett–Luce-based ranking objective.
    Predicted rankings are mapped to the discrete priority levels of the Choco solver before search begins, without modifying its propagation or search procedures. We evaluate the approach on XCSP3 instances and demonstrate that propagator ordering can have a substantial impact on solving time. On a test set of 698 instances, picked out as most difficult by problem class, the learned priorities increase the number of solved instances from 634 to 654, reduce median solver runtime by 27.93\%, and reduce aggregate penalized runtime (PAR1) by 12.65\% compared with Choco's default priorities. We further observe that improvements vary across problem classes and show that a simple policy selecting between learned and default priorities further increases the number of solved instances. These results demonstrate the potential of learning instance-dependent propagator priorities to improve constraint programming performance, particularly on difficult instances.}




\maketitle

\section{Introduction}

Constraint Programming (CP) solvers combine constraint propagation with search in order to solve combinatorial decision and optimization problems. Their performance often depends heavily on heuristic choices, such as variable selection, value selection and restart policies. 
Traditionally, these heuristics have been manually designed. 
In recent years, machine learning has increasingly been used to improve such heuristic choices, especially those related to branching decisions~\cite{ortizbayliss2013lvq, vandriel2021variable, chalumeau2021seapearl}. 
In this work, we investigate whether the same principle can be applied to a less commonly studied aspect of solver behavior: the order in which propagators are executed.

A propagator implements a filtering algorithm for a constraint, and during search many propagators may be scheduled for execution. Although propagation repeatedly applies scheduled propagators until a fixed-point is reached, execution order can still substantially affect runtime. Executing cheap propagators early incurs little immediate cost, but a more expensive propagator may justify its cost through stronger pruning that reduces downstream propagation and search. The relevant trade-off is therefore not simply about the cost of the next propagator, but about its cost relative to the work its pruning can eliminate. Because propagation is repeated throughout the search tree, these scheduling differences can accumulate into substantial differences in solving time. Most solvers therefore use priority-based scheduling schemes, although these priorities are typically fixed or hand-designed rather than adapted to the structure of a particular instance.

Learning propagator priorities presents a particular challenge: although the runtime of a solver under a given priority assignment can be measured, there is generally no direct supervision indicating which priorities should be assigned to individual propagators. The learning problem therefore requires relating the performance of complete priority assignments to predictions about their individual components. Learning from performance comparisons has previously been explored in algorithm configuration, including approaches based on pairwise comparisons and ranking models \cite{ijcai2015p49, weiss2025realtime}. More broadly, the proposed approach is related to learning-based algorithm selection and the adaptation of solver heuristics to individual problem instances \cite{nikolic2019portfolio, doolaard2025online}.

We investigate whether effective propagator priorities can be learned from static instance features and runtime measurements of alternative priority assignments. To the best of our knowledge, this is the first approach to learning static, instance-dependent propagator priorities. We formulate the problem as learning to rank propagators from comparisons of complete priority assignments.
At training time, for each training instance, we measure solver runtime under multiple randomized propagator priority orderings. These measurements are used to derive pairwise preferences between orderings, and a ranking model is trained to score propagators from extracted XCSP3 instance features. At solving time, the predicted scores are used to assign propagator priorities in the Choco constraint programming solver~\cite{prudhomme2014propagation} before search begins.

The experimental evaluation is performed on a filtered subset of XCSP3 instances. We deliberately use a difficulty-based training/validation/test split, to investigate generalization from easier to harder instances. The results show that propagator ordering can have a substantial effect on runtime, and that a learned ordering can improve performance on hard test instances compared with Choco's default ordering. The learned model solves more instances within the time limit and reduces aggregate solving runtime. We also show that the usefulness of the learned model varies across problem classes, motivating learning a simple policy that chooses between the learned ordering and the default ordering per problem class.

The main contributions of this work are:
\begin{itemize}

    \item We formulate the prediction of static, instance-dependent propagator priorities as a learning-to-rank problem, using runtime comparisons between complete priority assignments as supervision.
    \item We develop a self-attention-based ranking model that jointly scores the propagators present in an instance and integrate its predictions into the Choco solver without modifying its propagation or search procedures.
    \item We empirically demonstrate that propagator priority assignment can substantially affect solver performance and that learned priorities can improve the number of solved instances and aggregate runtime on difficult instances.
    \item We analyze how the benefits of learned priorities depend on instance difficulty and problem class, and demonstrate that a validation-based policy selecting between learned and default priorities can provide further improvements.    
\end{itemize}

The rest of the paper is organized as follows. Section~\ref{sec:related-work} discusses related work on learning-based heuristics, algorithm selection, learning-to-rank, and propagator scheduling. Section~\ref{sec:background} provides the necessary background on constraint propagation, propagator priorities, and set-based ranking. Section~\ref{sec:method} describes the proposed model and its architecture. Section~\ref{sec:experiments} presents the dataset, experimental setup, and evaluation results. Finally, Section~\ref{sec:conclusion} summarizes the findings and discusses possible directions for future work.

\section{Related work}
\label{sec:related-work}

Machine learning has been increasingly used to improve heuristic decisions in constraint programming and related combinatorial solvers. Much of this work focuses on decisions made during search, especially variable and value selection. For example, learning has been applied to variable-ordering heuristics, where the goal is to decide which variable should be branched on next, either by selecting among existing heuristics or by learning information used to initialize or adapt branching heuristics \cite{ortizbayliss2013lvq,vandriel2021variable}. Other work has considered learning branching decisions inside CP solvers using reinforcement learning~\cite{chalumeau2021seapearl}. These approaches share the motivation that solver heuristics are difficult to design manually and can have a large effect on performance. However, they mainly target branching decisions.

Our work is also related to algorithm selection and per-instance configuration. The algorithm-selection problem was formulated by Rice as the problem of selecting an algorithm for a given instance based on instance characteristics and performance data \cite{rice1976algorithm}. This idea has since been applied broadly to combinatorial solving, including constraint satisfaction, where instance features can be used to select among solvers or solving strategies. CPHydra, for example, uses case-based reasoning and CSP instance features to construct a solver portfolio for unseen instances \cite{omahony2008cphydra}, while Proteus performs hierarchical selection between different problem representations as well as different solvers using instance features \cite{hurley2014proteus}. 
Algorithm portfolios have also been investigated for SAT solving \cite{nikolic2013portfolio} and automated geometry theorem proving, where learned runtime predictions are used to select among alternative provers \cite{nikolic2019portfolio}. 
However, our main setting is more fine-grained than classical solver selection. Rather than selecting among different solvers, we modify an internal priority mechanism of a single CP solver.

The learning formulation used in this work is connected to learning-to-rank methods. Pairwise ranking methods learn a scoring function from comparisons between alternatives, where the model is trained to assign a higher score to the preferred item. RankNet is an early neural example of this approach, using pairwise preferences to learn a ranking function \cite{burges2005ranknet}. More recent neural ranking approaches have also considered ranking entire sets of alternatives jointly. SetRank, for example, uses self-attention to learn a permutation-invariant ranking function over sets of items \cite{pang2020setrank}. Our approach follows this set-based view of ranking, since for each constraint model, the alternatives to be ranked are the propagators that occur in that model. 

Finally, this work builds on previous research on propagation-engine design and propagator scheduling. Schulte and Stuckey study efficient propagation engines and, in particular, priority-based scheduling policies for deciding which propagator to execute next \cite{schulte2008propagation}. Choco likewise provides a propagation architecture in which propagators can be assigned priorities that influence their scheduling \cite{prudhomme2014propagation}. More recently, MetaJuLS applies meta-reinforcement learning to adaptive constraint-propagation scheduling, learning scheduling policies across constraint-programming benchmarks \cite{shihab2026metajuls}. In contrast, our approach predicts a static, instance-dependent ordering of propagators before search begins.

\section{Background}
\label{sec:background}

This section introduces the main concepts from constraint programming and machine learning needed to understand propagator ranking. We strive to make the concepts intuitively clear, without getting into technical details.

\subsection{Constraint solving and propagator scheduling}
\label{sec:background_cp}

A constraint satisfaction problem (CSP) consists of variables, the values each variable may take (its \emph{domain}), and constraints describing which combinations of values are allowed \cite{biere2021handbook}. For example, if two variables must be different and one is fixed to $3$, then the other cannot take the value $3$. Constraint programming (CP) solvers exploit such deductions to avoid examining every possible assignment. They are often employed in areas such as automated scheduling \cite{baptiste2001scheduling}, vehicle routing \cite{shaw1998vehicle}, employee rostering \cite{cheng1997nurse} and radio frequency assignment \cite{cabon1999radio}.

The process of removing values that cannot participate in a solution is called \emph{constraint propagation}. A \emph{propagator} implements an algorithm that inspects the current domains of variables that participate in a constraint and removes values that violate the constraint. Some propagators make simple, inexpensive deductions; others perform more involved reasoning and can eliminate many possibilities at once. Propagation is repeated because a domain reduction made by one propagator may enable another to remove additional values. This continues until no further reductions can be made (a \emph{fixpoint}) or a contradiction is found. If propagation alone does not solve the problem, the solver makes a choice during \emph{search}, such as assigning a value to a variable, and propagates again.

Running every propagator after every change would waste work. Instead, an \emph{event-driven propagation engine} keeps track of which propagators might be affected by changes to variable domains and places them in scheduling queues. When several propagators are waiting, the engine needs to decide which to execute first. Many solvers address this using \emph{priorities}: higher-priority propagators are processed before lower-priority ones. Choco uses seven priority levels, named \texttt{Unary}, \texttt{Binary}, \texttt{Ternary}, \texttt{Linear}, \texttt{Quadratic}, \texttt{Cubic}, and \texttt{VerySlow}, from highest to lowest priority \cite{prudhomme2014propagation}. 

The order matters because propagators differ both in their execution cost and in how much they can simplify the problem. A cheap propagator may quickly remove values and spare a more expensive one unnecessary work. Conversely, running a costly but powerful propagator early may simplify domains enough to save many later executions. Even when different execution orders eventually produce the same propagation fixpoint, they may require different amounts of computation to reach it. These differences can accumulate as propagation is repeated throughout search \cite{schulte2008propagation}.

\subsection{Machine learning and learning to rank}
\label{sec:background_rank}

\emph{Machine learning} aims to discover useful patterns from examples rather than specifying every decision rule by hand. A \emph{neural network} is a flexible parametrized model built from layers of simple computations: each layer transforms the information received from the previous one, and its parameters are adjusted during training so that the model's predictions better match the true target values which are available in the examples \cite{goodfellow2016deep}. The resulting network can learn relationships that would be difficult to express as fixed rules.

Neural networks work with numerical (often vectorial) representations of their inputs, which are transformed by layers to new numerical representations. Thus, the networks can produce vectorial representations of objects, often called \emph{embeddings}. Instead of treating two objects merely as equal or different, embeddings can capture similarities useful for the model's task. 

In \emph{learning to rank}, the goal is to place alternatives in a useful order rather than assign them independent categories. A ranking model typically gives each alternative a score enabling sorting of the alternatives by preference. Training information can come from \emph{pairwise preferences}: for example, observing that one ordering performs better than another tells the model which of the two should be favored, without requiring a known optimal ordering \cite{burges2005ranknet}.

In pairwise learning to rank, not all possible comparisons are necessarily equally useful for training. \emph{Pair mining} refers to the process of selecting which pairs of alternatives should be used as training examples \cite{wu2017sampling}. For example, comparisons between alternatives with very similar performance may provide a weaker or noisier learning signal, while comparisons with larger performance differences may be more informative. Different pair-mining
strategies can therefore be used to select or emphasize particular comparisons, potentially improving training efficiency and the quality
of the learned ranking.

Ranking a collection of objects introduces an important distinction between a \emph{sequence} and a \emph{set}. A sequence has meaningful positions, whereas a set does not: listing propagators as A, B, C rather than C, A, B should not change the score assigned to any propagator. A model with this property is called \emph{permutation equivariant}: reordering its inputs merely reorders the corresponding outputs. This is different from \emph{permutation invariance}, where the output itself stays unchanged, as when producing a single summary of a set \cite{zaheer2017deep, pang2020setrank}.

\emph{Self-attention} is a mechanism that lets a neural network represent each item using contextual information from the other items alongside it. Intuitively, every item can ``look at'' the others and learn which are most relevant when updating its own representation. \emph{Multi-head self-attention} extends this mechanism by using several attention mechanisms in parallel, allowing the model to capture different types of relationships between items and combine them into richer representations. A \emph{Transformer} combines self-attention with additional processing layers, allowing it to model interactions across an entire collection \cite{vaswani2017attention}. Although Transformers are often used for ordered sequences such as text, self-attention without positional information can also process sets while respecting their arbitrary input order \cite{lee2019settransformer}.

Finally, the \emph{Plackett--Luce model} connects scores of individual items to probabilities of complete rankings \cite{plackett1975analysis}. It assumes constructing a ranking one position at a time: first selecting an item from the whole collection, then one from the remaining items, and so on. Items with higher preference scores are more likely to be selected early, but less likely orderings are still possible. This gives a way to compare how strongly a scoring model favors two complete rankings, rather than comparing scores of individual items. A \emph{ranking loss} can then encourage the model to favor rankings better supported by the training examples.

\section{Method}
\label{sec:method}

This section describes our method for predicting propagator priorities. In short, given a constraint programming instance, the model assigns a score to each propagator present in its Choco representation. These scores define a ranking, which is converted into Choco's seven priority levels. Choco's propagation and search procedures are otherwise unchanged. 

\subsection{Problem formulation}
\label{sec:problem_formulation}

Let $x$ be a problem instance and let $P(x)=\{p_1,\ldots,p_n\}$ be the distinct propagators used in its Choco model. We also extract a vector of static instance features $z(x)\in\mathbb{R}^{d}$ of instance $x$, describing its properties such as the numbers and types of variables and constraints. These features are available before search and require no runtime observations from the instance being solved.

We seek a model which assigns each propagator a real-valued score,
\begin{equation}
    s_i=f_{\theta}\bigl(p_i,P(x),z(x)\bigr).
\end{equation}
such that an ordering of propagators induced by the scores yields (approximately) the shortest solver runtime among possible orderings.
Here, $\theta$ denotes the learned model parameters. Including the full set $P(x)$ means that a propagator's score may depend on which other propagators are present. We use the convention that \emph{lower scores indicate a preference for earlier execution}. Sorting the scores in ascending order produces the predicted propagator ranking.

The model operates over the vocabulary of propagators observed during training. Consequently, evaluation instances must contain only propagators from this vocabulary.

In order to find the desired model, we require:
\begin{itemize}
\item a model architecture which can take $p_i$, $P(x)$, and $z(x)$, equivariant in $P(x)$ and output a score for each propagator, 
\item a directly measurable runtime-based supervision for the model, and
\item a transformation from such supervision to propagator scores.
\end{itemize}

Direct supervision for individual propagator scores is generally unavailable, since the optimal priority assignment is unknown and the contribution of each propagator to overall solver performance cannot be measured independently. However, the relative quality of two complete priority assignments can be estimated by comparing their observed solver runtimes. The central learning challenge is therefore to infer scores for individual propagators from performance comparisons between complete priority assignments.

\subsection{Model architecture}
\label{sec:model_architecture}

\begin{figure}[htbp]
    \centering
    \includegraphics[
        width=0.8\linewidth,
        height=0.45\textheight,
        keepaspectratio
    ]{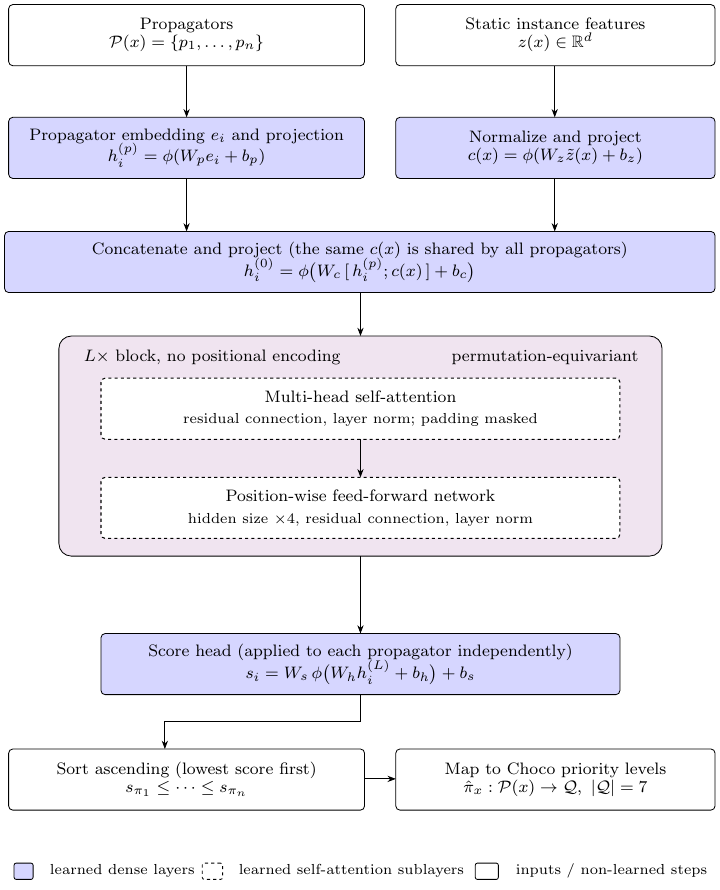}
    \caption{Architecture of the propagator ranking model.}
    \label{fig:architecture}
\end{figure}

Our architecture, based on the architecture of the SetRank model \cite{pang2020setrank} is shown in Figure~\ref{fig:architecture}. It has two inputs: the set of propagators in an instance and its static feature vector. The two inputs are first represented numerically, then combined and processed by self-attention blocks to produce a score for each propagator.

Each propagator $p_i$ has a learned embedding $e_i$, shared across instances in which that propagator occurs. The embedding is projected into the model's hidden space by:
\begin{equation}
    h_i^{(p)}=\phi(W_p e_i+b_p).
\end{equation}
In parallel, the static features are normalized using statistics from the training set and projected to an instance representation by:
\begin{equation}
    c(x)=\phi(W_z\widetilde{z}(x)+b_z).
\end{equation}
Here, $\phi$ denotes the activation function, $\widetilde{z}(x)$ denotes the normalized static feature vector and the weight matrices and bias vectors are learned during training.

The same instance representation $c(x)$ is combined with the representation of every propagator. After concatenation and projection, the input to the attention blocks is
\begin{equation}
    h_i^{(0)}=\phi\!\left(W_c[h_i^{(p)};c(x)]+b_c\right).
\end{equation}
Thus, the model starts with both propagator-specific information and information about the instance as a whole.

The combined representations pass through $L$ blocks of multi-head self-attention and position-wise feed-forward layers. As usual, each block uses residual connections and layer normalization; the feed-forward layer expands to four times the hidden dimension before projecting back. Self-attention allows a propagator's representation to reflect the other propagators present, rather than treating its scheduling preference as independent of them. Because propagators form a set, no positional encodings (otherwise usual in transformers) are added. When instances with different numbers of propagators are processed together, missing propagators are padded and masked.

Finally, the same score head is applied to each propagator representation after the last attention block:
\begin{equation}
    s_i=W_s\phi\!\left(W_hh_i^{(L)}+b_h\right)+b_s.
\end{equation}
The scores are sorted in ascending order and passed to the priority-assignment procedure in Section~\ref{sec:priority_assignment}. Although the score head is applied separately to each propagator, its inputs already incorporate information from the entire set.

\subsection{Training supervision}
\label{sec:training_supervision}

We obtain training examples by running Choco multiple times on each training instance, using different randomized assignments of propagators to priority levels. Each run records the assigned priorities and the resulting solver runtime. Priority assignments are converted to complete propagator orderings by sorting propagators by priority and applying a deterministic tie-break when propagators share a level.\footnote{Within each tied priority group, propagators are ordered using a deterministic pseudo-random key derived from the instance and run, avoiding a systematic alphabetical bias while remaining reproducible.} This tie-break provides a consistent representation for training; it does not imply that Choco executes tied propagators in that order.

Comparing runs on the \emph{same} instance gives a preference between their orderings: if one assignment solves the instance faster than another, its ordering is preferred. No optimal ordering needs to be known, and runtime measurements from different instances are not compared directly.

Runs that reach the time limit also provide useful information. With time limit $T$ and timeout penalty multiplier $\lambda\geq 1$, we define an effective runtime
\begin{equation}
    \tau(r)=
    \begin{cases}
        t(r), & t(r)<T,\\
        \lambda T, & t(r)\geq T,
    \end{cases}
\end{equation}
where $t(r)$ is the measured runtime of run $r$. 

For the sake of runtime comparisons, supervision will be given in the form of pairs of runtimes for various different propagator orderings on the same instance. 
Runs with equal effective runtime do not form a preference pair. Not every runtime difference is equally informative. For a faster run $r_f$ and a slower run $r_s$, we measure their relative difference as
\begin{equation}
    \Delta(r_f,r_s)=
    \frac{\tau(r_s)-\tau(r_f)}{\max(\tau(r_f),1)}.
\end{equation}
In the main training variant, we keep a pair only when this difference reaches a threshold $\delta$. Alternative pair-selection strategies are compared in Section~5.

\subsection{Training objective}
\label{sec:training_objective}

The supervision compares \emph{complete orderings}, whereas the network produces a score for each individual propagator. We bridge these two levels using the Plackett--Luce model~\cite{plackett1975analysis}. Since lower network scores indicate higher preference, we define the utility of propagator $p_i$ as $-s_i$.

For a complete ordering $o=(p_1,\ldots,p_n)$, its Plackett--Luce log-likelihood is
\begin{equation}
    \ell_{\theta}(o\mid x)=
    \sum_{t=1}^{n}
    \left[
        -s_t-
        \log\left(\sum_{j=t}^{n}\exp(-s_j)\right)
    \right].
\end{equation}
In effect, this assigns higher likelihood to orderings that provide lower scores for high-utility propagators, thus placing them earlier. Only propagators present in the instance enter this calculation.

For each retained pair, let $o_f$ and $o_s$ be the orderings associated with the faster and slower runs. We want the model to assign a higher likelihood to $o_f$. The margin loss we use is
\begin{equation}
    L_{\mathrm{pair}}(r_f,r_s)=
    \max\!\left(0,\,
        \gamma+\ell_{\theta}(o_s\mid x)
        -\ell_{\theta}(o_f\mid x)
    \right),
\end{equation}
where $\gamma\geq 0$ is a margin. The loss is zero when the faster ordering is favored by at least that margin; otherwise, it penalizes the model. 

Training minimizes the weighted mean loss over the set of retained pairs $B$:
\begin{equation}
    L(\theta)=
    \frac{\sum_{(r_f,r_s)\in B}
    w(r_f,r_s)L_{\mathrm{pair}}(r_f,r_s)}
    {\sum_{(r_f,r_s)\in B}w(r_f,r_s)}.
\end{equation}
where weights are defined using
$w(r_f,r_s)=\log(1+\Delta(r_f,r_s))$ 
to emphasize comparisons with clearer runtime differences without allowing very large ratios to dominate training.
This objective uses measured runtime preferences to train scores that induce useful propagator rankings, without requiring the model to predict runtime itself.

\subsection{Priority assignment}
\label{sec:priority_assignment}

At inference time, the predicted scores are sorted from lowest to highest. We then map this propagator ranking to Choco's seven priority levels, ordered from highest to lowest: \texttt{Unary}, \texttt{Binary}, \texttt{Ternary}, \texttt{Linear}, \texttt{Quadratic}, \texttt{Cubic}, and \texttt{VerySlow}.

If the instance contains at most seven distinct propagators, successive propagators in the predicted ranking receive successive priority levels. If there are more than seven propagators, the ranking must be compressed. Initially, each propagator forms its own group. We repeatedly merge the two adjacent groups with the smallest separation until seven groups remain. Here, separation denotes the smallest difference between the predicted scores of elements from adjacent groups. The obtained groups are then assigned to the seven Choco levels in ranked order. This preserves the overall ranking while allowing propagators with similar predicted preferences to share a level.

The resulting priorities are applied before the search begins. They remain fixed throughout the run; the solver's event handling, propagation algorithms, and search procedure are not modified.

\section{Experimental evaluation}
\label{sec:experiments}

This section evaluates whether learned propagator ordering can improve the performance of Choco on XCSP3 instances. We first describe the dataset and the collected runtime measurements, then analyze how strongly solver performance depends on propagator ordering. We describe the training setup. Finally, we compare the learned ordering model with Choco's default propagator ordering and discuss additional variants of the approach. The implementation of the proposed approach is publicly available at \url{https://github.com/Lojovic/ChocoRank}.

Experiments were performed on an x86\_64 machine with four AMD Opteron 6168 processors, for a total of 48 physical CPU cores and 94 GiB of RAM, running Ubuntu 24.04.1 LTS. Choco was executed using OpenJDK 21.0.12, with a 20 minute timeout per solver run. Test evaluations were parallelized over 8 worker processes, with each worker evaluating one instance at a time.

For runtime aggregation and timeout penalization, we use penalized average
runtime (PAR). Under PAR\(k\), a run that reaches the time limit \(T\) is
assigned runtime \(kT\), while completed runs retain their measured runtime.
Thus, PAR1 assigns a timeout the value \(T\), PAR2 assigns \(2T\), and PAR10
assigns \(10T\). Reporting multiple penalties distinguishes ordinary runtime
improvements from improvements due to timeout avoidance: larger values of
\(k\) place progressively greater weight on solving instances within the
time limit.

\subsection{Dataset description}

The experimental dataset was built from instances available in the XCSP3 \cite{boussemart2016xcsp3} instance collection.\footnote{https://www.xcsp.org/instances/} A total of 21,631 instances were downloaded from the \emph{Selected Instances} section, out of which 3,941 were selected for the purpose of our experiments by keeping only instances for which solving time was measured to be between 5 seconds and 20 minutes under a single run with randomized propagator priorities. The filtering is done to avoid instances that are too easy, for which runtime is dominated by measurement noise, as well as instances that are too hard, for which all measured runs would time out. Both of those scenarios would lead to data that is unsuitable for learning ordering preferences. Another reason for filtering is the duration of the experiment - running thousands of instances which hit the time out for each ordering could be very time consuming. Our filter is based on a single randomized-priority run, inclusion in the experimental subset can itself depend on the sampled priority assignment; we return to this limitation in Section~\ref{sec:conclusion}. However, we choose this way of filtering since it is not biased towards the default Choco ordering, or any other ordering for that matter. The selected instances span 99 different problem classes derived from instance names.

For each instance, a maximum of 20 runs were measured, each with different randomized propagator priorities, using a 20 minute timeout. The cap limits the computational cost of data collection while still sampling multiple alternative priority assignments for each instance. For instances with $n \le 3$ propagators, only the $n!$ available different propagator orderings were used. For each run, besides solver runtime, the present propagators were stored and their priorities were recorded as one of seven priority levels, with lower values corresponding to higher priority. The total number of measured runs is 62,994 and the number of distinct observed propagators is 62.

Model training and evaluation use a difficulty-based split performed separately within each problem class. Within each class, instances were ordered by difficulty, using the fraction of randomized orderings that timed out as the primary criterion and median solver runtime as the secondary criterion. The easiest 70\% of instances in each class were assigned to training, the next 15\% to validation, and the hardest 15\% to testing. Classes with fewer than seven instances were assigned entirely to the test set. This produced 2,672 training instances, 571 validation instances, and 698 test instances. This split was chosen to evaluate whether the learned model can generalize from easier and medium-difficulty instances to harder instances from the same problem classes. It deliberately introduces a difficulty shift between training and testing and should therefore be interpreted as an evaluation of generalization toward harder instances rather than as an  estimate of performance on the full instance distribution. This regime is particularly relevant here because scheduling improvements are most consequential on difficult instances, where a better priority assignment can substantially reduce runtime or prevent a timeout.

A total of 77 static instance features were extracted from each instance. The features used can be grouped into the following types:

\begin{itemize}
    \item constraint-level structural statistics, such as the number of constraints, constraint degrees, and domains of adjacent variables;
    \item domain/type statistics, such as information on Boolean and integer variables and constraints;
    \item global constraint category statistics, such as counts and ratios of global constraint families;
    \item objective-related statistics, such as objective domain and degree features;
    \item solving goal information, distinguishing satisfaction, minimization, and maximization instances;
    \item variable-level structural statistics, such as variable degrees, domains, and domain-degree products.
\end{itemize}

\begin{figure}[htbp]
    \centering
    \includegraphics[width=0.8\textwidth]{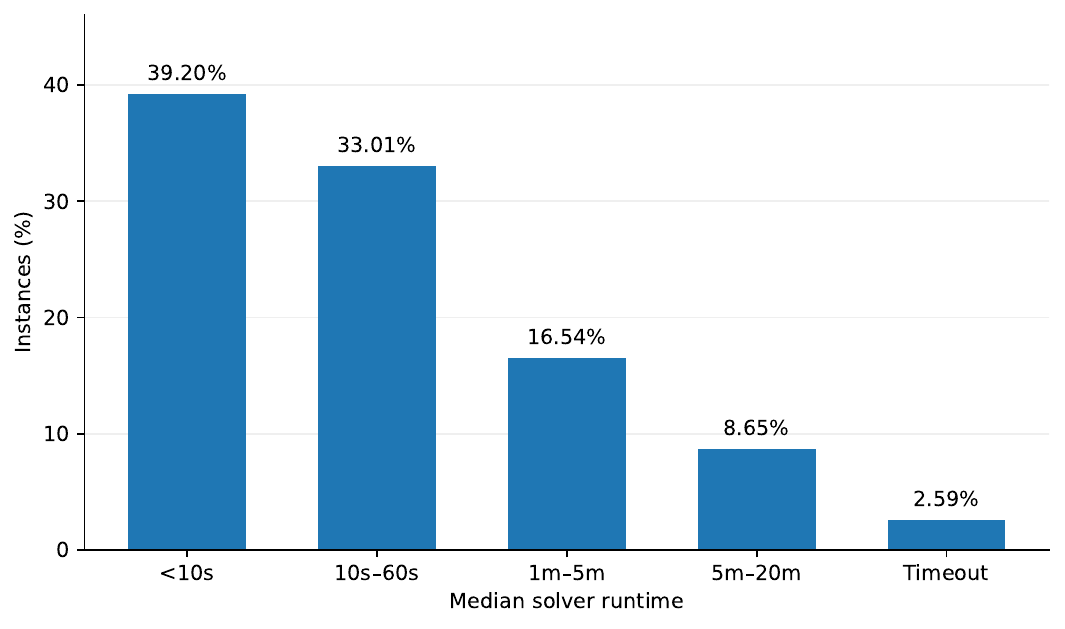}
    \caption{Distribution of instances by median solver runtime
    over measured randomized propagator orderings.}
    \label{fig:median-runtime-buckets}
\end{figure}

To characterize the difficulty of the selected instances, we grouped instances according to their median solver runtime over the measured randomized propagator orderings. The resulting distribution is shown in Figure~\ref{fig:median-runtime-buckets}.

The dataset contains a broad range of difficulties. Although a large fraction of instances have median runtimes below one minute, more than one quarter of the dataset has median runtime above one minute, and 102 instances have median runtime equal to timeout. This suggests that the filtered set contains many instances for which propagator ordering can have a meaningful runtime effect.

Timeout behavior provides an additional indication of ordering sensitivity. Most instances did not time out under any measured ordering, but a non-negligible subset timed out under some assignments. In particular, 295 instances had both a measured timeout and a measured successful run. Instances that timed out under every measured ordering do not yield pairwise training preferences because all measured runs receive the same timeout cost, but they remain useful for testing whether a learned ordering can solve difficult cases within the time limit.

\subsection{Impact of propagator ordering}

\begin{figure}[htbp]
    \centering
    \includegraphics[width=0.8\textwidth]{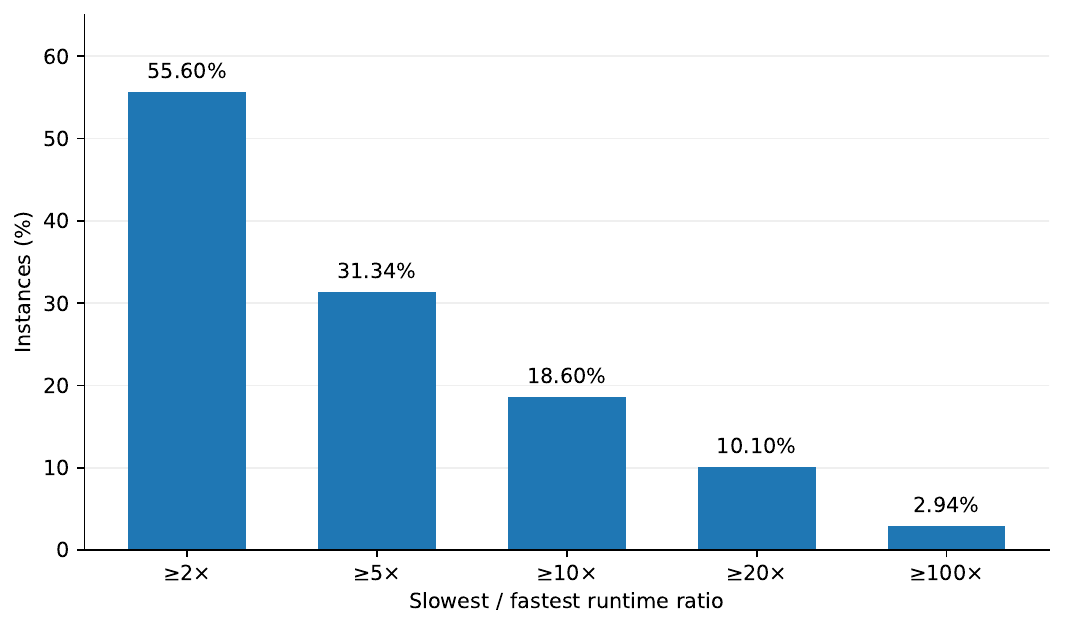}
    \caption{Impact of propagator ordering on solving time.
    Bars indicate the percentage of instances for which
    the ratio between the slowest and fastest randomized
    ordering exceeds the corresponding threshold.}
    \label{fig:ordering-impact-ratios}
\end{figure}

To measure the effect of propagator ordering independently of any learned model, we compared, for each instance, the fastest and slowest measured randomized ordering. Figure~\ref{fig:ordering-impact-ratios} shows how often this ratio exceeds different thresholds.

The median ratio between the slowest and fastest ordering was 2.29, and for 55.60\% of instances the slowest ordering was at least twice as slow as the fastest one. For 18.60\% of instances the difference was at least one order of magnitude. Moreover, in 295 instances, corresponding to 7.49\% of the dataset, the slowest measured ordering timed out while the fastest one solved the instance within the 20 minute limit. These results show that propagator ordering can substantially change solver behavior, both in runtime and in whether an instance is solved within the timeout, which motivates treating propagator ordering as a prediction problem. It should be noted, however, that this comparison is based only on the measured randomized orderings. Since at most 20 orderings were evaluated per instance, the fastest measured ordering is not guaranteed to be the optimal propagator ordering.

\subsection{Training configuration and model selection}
\label{subsec:training-configuration}

For each training run, we use pair mining to determine which pairs of measured propagator orderings are used as training examples. This selection can reduce noisy or uninformative comparisons between orderings whose runtimes differ only slightly. Different variations of this approach will be discussed in \ref{subsec:pairmining}.

The model was trained using Adam \cite{kingma2015adam}, with gradients clipped by global norm.
Hyperparameters were selected using Optuna \cite{akiba2019optuna} on the validation set. The
search space included the learning rate, Plackett--Luce margin, dropout
rate, pair-mining hyperparameters, pair-mining strategy, and timeout
penalty multiplier. Architectural hyperparameters, including the propagator
embedding dimension, hidden dimension, number of attention heads, number
of attention blocks, activation function, and gradient-clipping norm,
were fixed for a given hyperparameter search.

To avoid overfitting, as a usual practice, during training, different versions of the model are saved and evaluated on the validation set in search for the best one. 
Rather than selecting the version with the lowest training loss on the validation set, we select the one that minimizes 
a runtime-based validation criterion. The reason is that the training loss is a surrogate objective:
it measures consistency with preferences between measured orderings, whereas
the ultimate objective is to reduce solver runtime. We therefore select
the model version to retain using a criterion that is more directly aligned with 
solver performance. For a validation instance \(x\), let
\(R(x)\) denote its measured runs. For each run \(r\in R(x)\), the model
computes the Plackett--Luce log-likelihood
\[
    \ell_\theta(o_r\mid x).
\]
These values are normalized over the measured runs of the instance,
\[
    q_r
    =
    \frac{\exp(\ell_\theta(o_r\mid x))}
         {\sum_{r'\in R(x)}
          \exp(\ell_\theta(o_{r'}\mid x))}.
\]
The validation score for \(x\) is the expected effective runtime
\[
    V_\theta(x)
    =
    \sum_{r\in R(x)}q_r\tau(r).
\]
The overall validation objective is
\[
    V(\theta)
    =
    \sum_{x\in\mathcal{V}}V_\theta(x),
\]
where \(\mathcal{V}\) denotes the validation set. The model version minimizing
this objective is retained. 

Model training is implemented in TensorFlow. For solver-time inference, the trained parameters and model metadata are exported and evaluated using a separate NumPy implementation. This avoids the additional framework overhead of TensorFlow during solver execution and provides lower inference overhead for the relatively small model used in this work. It also allows the prediction pipeline to run without TensorFlow as a runtime dependency.

For an evaluation instance, the prediction pipeline determines the
propagators occurring in the Choco model, extracts and normalizes
the static XCSP3 features, evaluates the exported model, and converts the
resulting scores to Choco priority levels as described in
Section~\ref{sec:priority_assignment}. Choco then solves the instance
using the modified priorities.

\subsection{Model evaluation}

For the main evaluation, we use the model trained with threshold-based pair mining. We select this variant as the main model because it achieves the highest number of solved instances among the evaluated pair mining strategies, which is particularly relevant given our focus on difficult instances. In this setting, training pairs are created only from pairs of measured propagator orderings whose runtimes differ by at least a selected relative threshold. The purpose of this filtering is to focus training on ordering comparisons where the runtime difference is large enough to provide a meaningful learning signal. The threshold used for pair mining was selected on the validation set together with the remaining hyperparameters.

The final model was evaluated on the test set against Choco's default propagator ordering. We report both walltime and solver runtime. Solver runtime denotes the time spent by Choco during solving, while walltime denotes the total end-to-end runtime of the execution. For the learned model, walltime therefore also includes the overhead of collecting the propagators, extracting instance features, running model inference, and assigning the predicted priorities before solving. The results are shown in Table~\ref{tab:main-model-results}.

 \begin{table}[htbp]
\caption{Comparison between Choco's default propagator ordering and the learned threshold pair-mining model on the hard test set.}
\label{tab:main-model-results}
\centering
\begin{tabular}{lrrr}
\hline
Metric & Default & Model & Improvement \\
\hline
Solved instances & 634/698 & 654/698 & +20 \\
Median walltime & 168.37s & 143.14s & 14.98\% \\
Median solver runtime & 167.88s & 120.99s & 27.93\% \\
Aggregate walltime, PAR1 & 235,653s & 210,410s & 10.71\% \\
Aggregate solver runtime, PAR1 & 235,260s & 205,510s & 12.65\% \\
Aggregate walltime, PAR2 & 312,453s & 263,210s & 15.76\% \\
Aggregate solver runtime, PAR2 & 312,060s & 258,310s & 17.22\% \\
Aggregate walltime, PAR10 & 926,853s & 685,610s & 26.03\% \\
Aggregate solver runtime, PAR10 & 926,460s & 680,710s & 26.53\% \\
\hline
\end{tabular}

\end{table}

The learned model solved 20 more instances than the default ordering. It also reduced median walltime by 14.98\% and median solver runtime by 27.93\%. The improvement is larger under PAR2 and PAR10, as these metrics
assign greater penalties to timeouts, thereby amplifying the
performance of the method which solves more instances.

To better understand where the learned ordering is useful, we grouped test instances by the solver runtime of Choco's default ordering. The results are shown in Table~\ref{tab:results-by-difficulty}.

\begin{table}[htbp]
\caption{Model performance by instance difficulty, where difficulty is measured by Choco's default solver runtime. Runtime values report median solver runtime in each group.}
\label{tab:results-by-difficulty}
\centering
\begin{tabular}{lrrrl}
\hline
Def. runtime & $n$ & Solved def. & Solved model & Runtime def./model \\
\hline
$<10$s & 95 & 95 & 95 & 6.11s / 6.63s \\
10s--60s & 143 & 143 & 141 & 29.95s / 35.32s \\
1m--5m & 204 & 204 & 201 & 160.21s / 120.02s \\
5m--20m & 192 & 192 & 179 & 566.09s / 447.09s \\
Timeout & 64 & 0 & 38 & 1200.00s / 1049.68s \\
\hline
\end{tabular}

\end{table}

The learned model is not uniformly beneficial across all difficulty ranges. For instances solved by the default ordering in under one minute, the predicted priorities provide little benefit and can increase solver runtime. The model becomes more useful on harder instances: for instances solved by the default ordering in 5 to 20 minutes, it reduces median solver runtime from 566.09s to 447.09s. The strongest effect appears on instances where the default ordering timed out, where the learned ordering solves 38 out of 64 instances. In addition to differences caused by the predicted priorities themselves, applying the model introduces a median end-to-end overhead of 1.72s, measured as the difference between model walltime and Choco solver runtime on instances solved by the model. This overhead is particularly relevant for easy instances, where the potential runtime reduction is comparatively small.

To examine whether the learned ordering is equally useful across problem classes, we grouped test instances by the problem class derived from the instance name. Table~\ref{tab:results-by-class} shows the classes with the largest positive and negative aggregate PAR1 solver-runtime differences, considering only classes with at least five test instances.

\begin{table}[htbp]
\caption{Problem classes with the largest improvements and regressions under the learned threshold pair-mining model. PAR1 diff is the aggregate Choco solver runtime of the default ordering minus that of the learned ordering, with timeouts counted as 1200s. Median values report Choco solver runtime.}
\label{tab:results-by-class}
\centering
\begin{tabular}{lrrrr}
\hline
Class & $n$ & Solved def./model & PAR1 diff & Median def./model \\
\hline
\multicolumn{5}{c}{Largest improvements} \\
\hline
Crossword & 47 & 38 / 46 & 11,693s & 639.1s / 302.7s \\
Subisomorphism & 96 & 78 / 87 & 9,540s & 297.5s / 215.6s \\
qwh & 5 & 0 / 4 & 3,206s & 1200.0s / 376.9s \\
Wwtpp & 52 & 49 / 48 & 2,467s & 172.5s / 102.9s \\
SocialGolfers & 26 & 23 / 24 & 2,369s & 72.0s / 65.9s \\
PrizeCollecting & 5 & 4 / 5 & 1,256s & 1084.8s / 835.2s \\
QueenAttacking & 6 & 4 / 5 & 1,123s & 19.0s / 34.0s \\
CarSequencing & 9 & 7 / 8 & 1,017s & 302.7s / 294.0s \\
Pb & 25 & 24 / 25 & 881s & 275.6s / 201.3s \\
Bibd & 10 & 9 / 8 & 725s & 665.5s / 512.0s \\
\hline
\multicolumn{5}{c}{Largest regressions} \\
\hline
TravellingSalesman & 6 & 6 / 6 & -691s & 236.2s / 410.9s \\
SchurrLemma & 5 & 5 / 5 & -588s & 251.8s / 432.6s \\
Fischer & 9 & 9 / 7 & -472s & 374.7s / 370.5s \\
Rcpsp & 46 & 46 / 46 & -199s & 106.8s / 113.0s \\
SuperSadeh & 6 & 6 / 6 & -163s & 73.0s / 85.6s \\
bdd & 10 & 10 / 10 & -128s & 188.3s / 196.3s \\
MultiKnapsack & 6 & 6 / 6 & -106s & 17.6s / 18.0s \\
GracefulGraph & 6 & 6 / 6 & -97s & 21.8s / 46.7s \\
PropStress & 5 & 5 / 5 & -77s & 13.1s / 16.0s \\
reg & 14 & 14 / 14 & -38s & 60.0s / 63.3s \\
\hline
\end{tabular}

\end{table}

The results show that the learned ordering is not uniformly beneficial across problem classes. The largest gains appear for classes such as Crossword and Subisomorphism, where the model both solves more instances and substantially reduces aggregate solver runtime. On the other hand, several classes exhibit regressions, including TravellingSalesman, SchurrLemma, and Fischer. This suggests that the usefulness of learned propagator ordering depends strongly on the problem class. As for the problem classes with fewer than seven instances, which were placed entirely in the test set, the results were mixed: the model solved one additional instance and achieved a 2.1\% lower aggregate PAR1 walltime, but it was slower on most individual instances with the median walltime being 16.08s versus the median walltime of 10.86s for the default ordering.

Motivated by the observed variation across problem classes, we additionally
evaluate a validation-based gated policy. Let \(c(x)\) denote the problem
class of instance \(x\). For each class \(c\) represented in the validation
set, we evaluate both the learned priority model and Choco's default
priorities on the validation instances belonging to \(c\). Let
\(A_{\mathrm{model}}(c)\) and \(A_{\mathrm{default}}(c)\) denote their
aggregate PAR1 walltimes, and let \(N_{\mathrm{model}}(c)\) and
\(N_{\mathrm{default}}(c)\) denote their numbers of solved instances.

The policy selects the learned model for class \(c\) if
\[
    A_{\mathrm{model}}(c)
    <
    A_{\mathrm{default}}(c)
\]
and
\[
    N_{\mathrm{model}}(c)
    \geq
    N_{\mathrm{default}}(c).
\]
Otherwise, Choco's default priorities are selected. Classes not represented
in the validation set fall back to the learned model. The resulting policy
assigned 19 observed problem classes to the learned model and 35 to the
default priorities.

\begin{table}[htbp]
\caption{Comparison between the ungated threshold model and the validation-based gated policy on the test set. The gated policy chooses between the learned threshold model and Choco's default ordering based on validation PAR1 walltime for each problem class. Positive percentages indicate improvement of the gated policy over the ungated model.}
\label{tab:gated-policy-results}
\centering
\begin{tabular}{lrrr}
\hline
Metric & Ungated model & Gated policy & Change \\
\hline
Solved instances & 654/698 & 657/698 & +3 \\
Median walltime & 143.14s & 129.93s & 9.24\% \\
Median solver runtime & 120.99s & 127.83s & -5.66\% \\
Aggregate walltime, PAR1 & 210,410s & 204,399s & 2.86\% \\
Aggregate solver runtime, PAR1 & 205,510s & 203,252s & 1.10\% \\
Aggregate walltime, PAR2 & 263,210s & 253,599s & 3.65\% \\
Aggregate solver runtime, PAR2 & 258,310s & 252,452s & 2.27\% \\
Aggregate walltime, PAR10 & 685,610s & 647,199s & 5.60\% \\
Aggregate solver runtime, PAR10 & 680,710s & 646,052s & 5.09\% \\
\hline
\end{tabular}

\end{table}

The gated policy improves over the ungated threshold model on most metrics. It solves three additional instances and reduces aggregate PAR1 walltime by 2.86\%. Median solver runtime is slightly worse for the gated policy. This is expected because the policy can reduce walltime by avoiding model overhead on classes assigned to the default ordering, while the default ordering may still have slightly worse pure solver runtime than the learned ordering on some of those instances. The results support the observation that learned propagator ordering is problem-class dependent, and that validation data can be used to avoid applying the model on classes where it is unlikely to help. 

One possible reason why learned ordering is more useful for some classes than others is the presence of adaptive propagators, whose behavior can depend on the previous solving trajectory. In such cases, changing the order in which propagators are executed may affect the later state of the search more strongly than for other propagators. A more detailed analysis of this effect is left for future work.

\subsection{Alternative pair mining strategies}
\label{subsec:pairmining}

We compared the threshold-based pair-mining strategy used in the main evaluation with three alternative strategies. In all cases, a training pair consists of two measured propagator orderings for the same instance, where the faster ordering is preferred over the slower one. Pair construction uses effective runtimes, where timeout runs are replaced by a penalized runtime according to the timeout penalty multiplier selected during training. The threshold strategy keeps only pairs whose relative runtime difference exceeds a selected threshold. The weighted strategy keeps pairs probabilistically: pairs with larger relative runtime differences are more likely to be selected, while a minimum keep probability is imposed by a floor hyperparameter. The top-K/bottom-K strategy ranks the measured orderings for each instance and constructs pairs only between the $K$ fastest and $K$ slowest orderings, where $K$ is a model hyperparameter. Finally, the all-pairs strategy performs no pair mining and keeps every pair with unequal effective runtime. All variants use the same model architecture and the same difficulty-based train/validation/test split.

\begin{table}[htbp]
\caption{Comparison of alternative pair-mining strategies. Aggregate times use PAR1, with timeouts counted as 1200s.}
\label{tab:pair-mining-comparison}
\centering
\begin{tabular}{lrrrrr}
\hline
Variant & Solved & Agg. walltime & Agg. solver & Med. walltime & Med. solver \\
\hline
Threshold & 654/698 & 210,410s & 205,510s & 143.14s & 120.99s \\
Weighted & 650/698 & 212,447s & 207,596s & 139.04s & 121.87s \\
Top-K/bottom-K & 647/698 & 210,377s & 205,435s & 137.20s & 119.80s \\
All pairs & 648/698 & 211,887s & 207,083s & 148.61s & 126.40s \\
\hline
\end{tabular}

\end{table}

The variants produce similar results, suggesting that performance is relatively insensitive to the exact pair-mining strategy. Threshold-based pair mining solves the largest number of instances, while top-K/bottom-K achieves slightly lower aggregate and median runtimes but solves fewer instances. The competitive all-pairs result further suggests that removing small-difference pairs is not essential for learning useful propagator orderings. Since the main goal is robustness on difficult instances, we use the threshold model as the central model because it solves the largest number of test instances.

\section{Conclusion}
\label{sec:conclusion}

This work investigated whether propagator priorities in constraint programming can be predicted from static instance characteristics. We formulated the task as set-based learning to rank: propagators are scored jointly from static XCSP3 features, and the resulting ranking is mapped to Choco's existing priority levels before search. Training supervision is derived from runtime measurements under randomized propagator priority assignments, allowing priorities to be learned without modifying Choco's subsequent propagation or search procedure.

The experiments first demonstrate that propagator priority assignment can have a substantial effect on solver performance. Across the collected dataset, the median ratio between the slowest and fastest measured randomized ordering was \(2.29\), while for \(55.60\%\) of instances the slowest measured ordering was at least twice as slow as the fastest. Moreover, for 295 instances, one measured ordering timed out while another solved the same instance within the time limit. These results indicate that propagator scheduling constitutes a meaningful source of performance variation and that fixed priorities can leave considerable room for instance-dependent improvement.

On the difficulty-based test set, the learned ordering improved over Choco's default priorities overall. The main model solved \(654\) of \(698\) instances compared with \(634\) for the default ordering, reduced median solver runtime by \(27.93\%\), and reduced aggregate PAR1 solver runtime by \(12.65\%\). The improvements were particularly pronounced on difficult instances and under larger timeout penalties. At the same time, the learned ordering was not uniformly beneficial. Performance varied substantially across problem classes, and the model could regress on instances for which Choco's default ordering was already effective. A validation-based gated policy that selected between the learned and default priorities according to problem class further improved the solved count to \(657\) instances and reduced aggregate runtime. This suggests that learned and manually designed priority policies have complementary strengths rather than one policy being uniformly preferable.

Several limitations provide directions for future work. Because initial dataset filtering is based on a single randomized-priority run, inclusion in the experimental subset can itself depend on the sampled priority assignment. The current model is also restricted to propagators represented in its training vocabulary, motivating representations that can generalize to previously unseen propagators. Furthermore, the observed variation between problem classes suggests that model selection or gating could itself be learned from instance characteristics rather than relying on predefined problem-class labels. Finally, a more detailed analysis of individual propagators and model performance on instances in which they appear could yield different gating strategies based on propagators present in an instance.

Overall, the results show that propagator priority assignment is a consequential heuristic choice and that useful priority information can be learned from static instance properties. The gains are not uniform, but the improvements on difficult cases motivate further investigation of learned propagation scheduling as a complementary component of constraint programming solvers.
\backmatter


\section*{Declarations}

\subsection*{Funding}

Both authors are partially supported by the Ministry of  Science, Technological Development and Innovation, Republic of Serbia, through the project 451-03-33/2026-03/200104.

\subsection*{Competing interests}
 The authors have no competing interests to declare that are relevant to the content of this article.

\subsection*{Author contributions}

Both authors worked on the idea of the paper. The first author implemented the approach, conducted experiments, and wrote the first draft of the manuscript. The second author provided guidance on all aspects of the work, especially the deep learning approach, and worked on improving the manuscript.

\bibliography{references}

\end{document}